\pdfoutput=1

\documentclass[11pt]{article}
\usepackage{color, colortbl}
\definecolor{Gray}{gray}{0.9}
\usepackage{graphicx}

\usepackage{amsmath}
\usepackage{makecell}

\usepackage{booktabs}
\usepackage{adjustbox}

\usepackage[final]{acl}

\usepackage{times}
\usepackage{latexsym}

\usepackage[T1]{fontenc}
\usepackage[utf8]{inputenc}

\usepackage{microtype}
\usepackage{inconsolata}

\usepackage{todonotes}

\title{Mitigating Hallucinations in Large Vision-Language Models (LVLMs) \\ via Language-Contrastive Decoding (LCD)}

\author{Avshalom Manevich \\
  Bar Ilan University \\
  \texttt{avshalomman@gmail.com} \\\And
  Reut Tsarfaty \\
  Bar Ilan University \\
\texttt{reut.tsarfaty@biu.ac.il} \\}

\begin{document}
\maketitle
\begin{abstract}
Large Vision-Language Models (LVLMs) are an extension of Large Language Models (LLMs) that facilitate processing both image and text inputs, expanding AI capabilities. However, LVLMs struggle with object hallucinations due to their reliance on text cues and learned object co-occurrence biases. While most research quantifies these hallucinations, mitigation strategies are still lacking. Our study introduces a Language Contrastive Decoding (LCD) algorithm that adjusts LVLM outputs based on LLM distribution confidence levels, effectively reducing object hallucinations. We demonstrate the advantages of LCD in leading LVLMs, showing up to 4\% improvement in POPE F1 scores and up to 36\% reduction in CHAIR scores on the COCO validation set, while also improving captioning quality scores. Our method effectively improves LVLMs without needing complex post-processing or retraining, and is easily applicable to different models. Our findings highlight the potential of further exploration of LVLM-specific decoding algorithms for improved multimodal performance. 
\end{abstract}

\begin{figure}[!htb]
\centering
\includegraphics[scale=0.42]{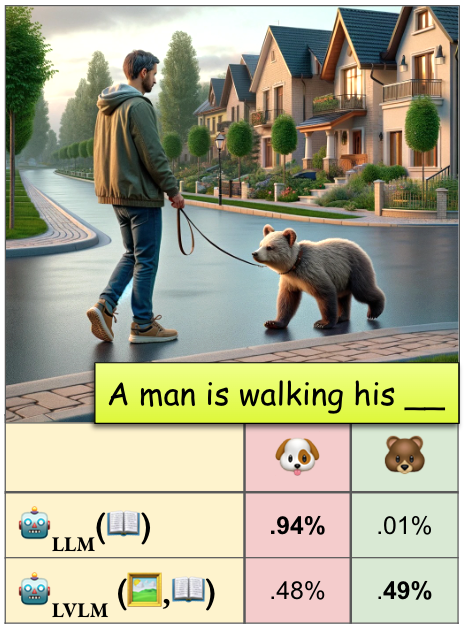}
\caption{An illustration of LLM vs. LVLM token probabilities given an image and a text prefix mid-generation. According to the LLM, the word "dog" is much more likely to appear next. LCD dynamically contrasts these probabilities to mitigate language biases in LVLM outputs.}
\label{fig:ill1}
\end{figure}

\section{Introduction}

Large Vision-Language Models (LVLMs) are a multimodal extension of Large Language Models (LLMs), transforming textual prompts and image inputs into text. However, they frequently produce object hallucinations, where absent objects are mentioned in the output \citep{Li-hallucination-2023, lovenia2023negative}.

While hallucination-mitigation techniques in LLMs are actively researched, specific strategies for LVLMs are less developed. 
Current methods involve model-specific adjustments, additional training, or auxiliary models for post-hoc correction, and are often proven inefficient, costly, or limited by training data and model biases \citep{wang2023mitigating, zhou2023analyzing, gunjal2023detecting, yin2023woodpecker}. 
Conversely, LVLM hallucination evaluation has seen progress with object hallucination  benchmarks like NOPE \citep{lovenia2023negative} and POPE \citep{Li-hallucination-2023}, and recent works that aim for more holistic LVLM hallucination evaluation such as FaithScore \citep{faithscore} and HallusionBench \citep{guan2023hallusionbench}. 

A key reason for LVLM hallucinations is their tendency to over-rely on linguistic information, as was first observed by \citet{guan2023hallusionbench}.
Based on this insight, we propose to intervene in the LVLM decoding phase so that model outputs are less informed by language biases.
Specifically, we propose to use Contrastive Decoding \citep{li2023contrastive, obrien2023contrastive} to alter LVLM output probabilities with respect to the internal LLM's probabilities, guided by a dynamic weighting mechanism based on the LLM distribution's entropy.

Our experiments show that our proposed method, Language Contrastive Decoding (LCD), improves hallucination scores on POPE \citep{Li-hallucination-2023} and CHAIR \citep{objectHallucination} on InstructBLIP variants based on Vicuna and Flan-T5 \citep{dai2023instructblip}, LLAVA 1.5 \citep{liu2023improvedllava} and mPLUG-Owl2 \citep{ye2023mplugowl2}. We asses LCD's overall generation quality by reporting captioning metrics and conducting a GPT4-V \cite{openai2023gpt4} assisted evaluation. LCD, as a decoding strategy, can be applied to other models without additional training or output modifications, emphasizing its utility for broader LVLM use. 

The contributions of this paper are thus manifold. First,  we introduce a novel decoding method tailored for LVLMs to mitigate object hallucinations. Next, we present a dynamic weighting strategy based on entropy which is applicable in various CD scenarios. Finally, we share our code to encourage further research into LVLM-specific decoding strategies, a promising avenue for future research.

\section{Motivation and Background}
The integration of vision capabilities into LLMs has led to the development of Large Vision-Language Models, merging LLMs' textual understanding with vision-text encoders. This trend towards multimodal systems is exemplified in commercial platforms such as GPT4-V  \citep{openai2023gpt4} and Google's Gemini \citep{geminiteam2023gemini}.

\paragraph{Large Vision-Language Models} combine LLMs and vision-text encoders to generate text from textual prompts and visual inputs. An LVLM generally comprises three main components: a vision-text encoder like CLIP \citep{radford2021learning}, an LLM such as LLAMA \citep{touvron2023llama} or Flan-T5 \citep{chung2022scaling}, and a cross-modal alignment module linking the vision-text encoder output with the LLM. 

Initially, LVLMs were fine-tuned for specific tasks \citep{li2022blip, wang2022git}. However, advancements in LLMs have led to a shift towards general-purpose, instruction-tuned LVLMs. These models are designed to handle a wide range of tasks based on instructions, making them more versatile.
Despite these advancements, LVLMs grapple with hallucinations of different types.

\paragraph{LVLMs Hallucinations and their Mitigation} 
Hallucinations in LVLMs, particularly object hallucinations where nonexistent entities are mentioned, are often attributed to LVLMs' reliance on spurious correlations and language biases, as demonstrated by \citet{li2023evaluating} and \citet{zhou2023analyzing}. Moreover, \citet{guan2023hallusionbench} highlight LVLMs' tendency to prioritize language over visual data, leading to hallucinations.

Mitigation strategies proposed by \citet{gunjal2023detecting} and \citet{wang2023mitigating} involve further model training with augmented datasets or reward models. \citet{zhou2023analyzing, yin2023woodpecker} developed auxiliary models to correct outputs post-generation. These solutions often require dataset-specific work or additional model training, potentially leading to overfitting or new biases, and are not easily transferable across LVLMs.

In a concurrent work, \citet{leng2023mitigating} develop an LVLM-specific decoding algorithm for mitigating hallucinations, using a noisy copy of the input image as a contrastive input. While their approach uses visual noise to guide the decoding process, LCD leverages the language modality to mitigate hallucinations. These approaches are orthogonal and can potentially be combined into a unified Language-Visual contrastive decoding algorithm, a direction we leave for future work.
\footnote{\citet{favero2024multimodal} propose a method with a high resemblance to ours, however, our work predates theirs. https://openreview.net/forum?id=aReb-02mhR }

\section{Language Contrastive Decoding (LCD)}
Before presenting LCD, we  briefly introduce the essentials of  decoding in LVLMs~\ref{sec:essential}, followed  by our formal proposal~\ref{sec:proposal} and research hypothesis~\ref{sec:hypothesis}.
\subsection{Decoding Techniques and Contrastive Decoding: Essential Preliminaries}\label{sec:essential}

Decoding in auto-regressive generative models is the stage that transforms an input representation into a sequence of output tokens. In LVLMs, this process involves a model $M$, an image $I$, a textual prompt $X$, and a particular timestamp $t$ during generation. It can be described as a series of selections from the model's probability distribution, producing a token sequence $T$, as formalized in eq.~(\ref{eq:1}).
\begin{equation}
T_t \sim P(\cdot | I, X, T_{<t}; M)
\label{eq:1}
\end{equation}

Greedy decoding, selecting the most probable token at each step (or the top $k$ tokens in a beam search with beam size $k$), is the simplest approach. 
However, high likelihood sequences do not necessarily align with human preferences, leading to the ``likelihood trap'' \citep{zhang-etal-2021-trading}. This has led to the use of sampling-based methods, such as top-k sampling, nucleus sampling \citep{holtzman2020curious}, and locally typical sampling \citep{meister2023locally}, which either truncate the set of candidate tokens or adjust the model's distribution, e.g.\ through temperature scaling.

Contrastive Decoding (CD) has been introduced for LLMs as a method to penalize the outputs of an expert model with those from a less powerful model \citep{li2023contrastive}. CD can be applied to any two probability distributions with the same support and has been adapted as a sampling strategy, improving various text generation tasks \citep{obrien2023contrastive, chuang2023dola, sennrich2024mitigating}.
CD uses both truncation and reshaping of probability distributions. The truncation phase ("adaptive plausibility") is described by eq.~(\ref{eq:2}), where $\alpha$ is a hyper-parameter, $\mathcal{V}$ and $\mathcal{V}t'$ are the original and truncated token vocabularies at time $t$, and $P$ is the conditional distribution on the prefix $T_{<t}$.
\begin{equation} \mathcal{V}t' = \{ v \in \mathcal{V} : P(v|T_{<t}) \geq \alpha \max_{w} P(w|T_{<t}) \} \label{eq:2} \end{equation}

Finally, the formula for CD, as suggested by \citet{obrien2023contrastive}, given here generally for two conditional distributions $P$ and $P'$ on variable $x$ with the same support, conditioned on a prefix sequence $X$ is presented in eq.~(\ref{eq:3}).

{\small  
\begin{align} \nonumber & CD{t}(x, X, P, P') =\\ & \begin{cases} (1 + \beta)\log P(x|X) - \beta \log P'(x|X), & \text{if } x \in Vt' \\ -\infty, & \text{otherwise} \end{cases} \label{eq:3} \end{align}
}%

$\beta$ is a fixed weight hyper-parameter. Our  proposed method, detailed shortly, alters CD by introducing an entropy-based dynamic weighting scheme.

\subsection{Proposed Method}\label{sec:proposal} Our intuition, based on previous findings by \cite{guan2023hallusionbench, objectHallucination, Li-hallucination-2023}, is that an LVLM can be "misled" by its constituent LLM during the generation process. 

Consider for example an LVLM that is describing an image (see illustration \ref{fig:ill1}). Mid-generation, given the text "An image of a man walking his," it may predict "dog" due to language biases, even if it is a bear  that is actually shown. A `plain' LLM, without seeing the image, reinforces these biases by highly rating ``dog''. Our method builds on this insight to guide an LVLM towards more accurate predictions using Contrastive Decoding.

Our method operates as follows: At each generation step $t$, for each token $x$, we first determine the next-token probabilities from the LVLM, $P_{LVLM}$, based on the current token sequence $T_{<t}$, text $X$, and image $I$. We then obtain a second distribution, $P_{LLM}$, by inputting all data except the image into the LLM. The LLM's conditional entropy $\mathrm{H}_{LLM}$ informs the dynamic weight as per eq.\eqref{eq:4}. We then adjust token $x$'s logits using the LCD formula in eq. \eqref{eq:5}.
\begin{align}
\beta_t = \frac{\beta}{\mathrm{H}_{LLM}(x|X, T_{<t})}
\label{eq:4}
\end{align}
\begin{equation}
\begin{aligned}
LCD_{t}(x, T_{<t}, I, P_{LVLM}, P_{LLM}) &= \\
&\hspace{-9em} (1 + \beta_{t})\log P_{LVLM}(x|I, X, T_{<t}) \\
&\hspace{-10em} - \beta_{t} \log P_{LLM}(x|X, T_{<t})
\end{aligned}
\label{eq:5}
\end{equation}
In our experiments, we generate text completions by sampling from the next token probabilities, which are obtained by applying the softmax function to the logits produced by the LCD algorithm.

\begin{table*}[t]

\centering\scalebox{0.88}{
\begin{tabular}{ll|lll|ll|ll}
\hline
\small\textbf{Model} & \small\textbf{Method} & \small\textbf{METEOR}$\uparrow$ & \small\textbf{WMD}$\uparrow$ & \small\textbf{ROUGE\textsubscript{L}}$\uparrow$ & 
\small\textbf{Acc}$\uparrow$ &
\small\textbf{Det}$\uparrow$  & \small\textbf{CHAIRs}$\downarrow$ & \small\textbf{CHAIRi}$\downarrow$ \\ \hline
InstructBLIP$_\text{F}$ & Baseline & .157 & .367 & .161 & 4.92 & 4.02 & .662 & .146 \\
 & LCD & \textbf{.159} & \textbf{.370} & \textbf{.168} & \textbf{5.4} & 4.01 & \textbf{.566} & \textbf{.131} \\
\hline
InstructBLIP$_\text{V}$ & Baseline & .178 & .423 & .291 & 3.7 & 3.51 &  .274 & .126 \\
 & LCD & \textbf{.199} & \textbf{.48} & \textbf{.38} & \textbf{4.59} & \textbf{3.83} & \textbf{.174} & \textbf{.107} \\
\hline
LLAVA 1.5 & Baseline & .163 & \textbf{.357} & .169 & 4.77 & 4.56 & .672 & .182 \\
 & LCD & \textbf{.171} & .352 & \textbf{.181} & \textbf{5.39} & 4.54 & \textbf{.610} & \textbf{.161} \\
 \hline
mPLUG-Owl2 & Baseline & .162 & .357 & .163 & 4.68 & 4.7 & .660 & .19 \\
 & LCD & \textbf{.177} & \textbf{.372} & \textbf{.184} & \textbf{5.11} & 4.69 & \textbf{.614} & \textbf{.145} \\
 \hline
\end{tabular}}
\parbox{\textwidth}{\caption{Image Description results. \textit{F} and \textit{V} stand for the Flan-T5 and Vicuna. Acc and Det are mean GPT4-V scores for Accuracy and Detailedness. METEOR, WMD and ROUGE$_\text{L}$ are popular captioning metrics \citep{kusner2015doc, banerjee-lavie-2005-meteor, lin-2004-rouge}. $\uparrow$ means `higher is better'. $\downarrow$ means `lower is better'.
}
\label{table:description_results}
}
\end{table*}

\subsection{Research Hypothesis}\label{sec:hypothesis}
Our hypothesis is that contrasting LVLM outputs with LLM outputs conditioned only on the textual data, can mitigate language biases, therefore reducing hallucinations in LVLMs.

\section{Experiments and Results}

We set out to assess the effect of LCD on object hallucinations in LVLM outputs against popular decoding settings. Additionally, we verify that LCD does not degrade output quality. To this end, we asses LCD on the POPE benchmark \citep{Li-hallucination-2023}, and on an image detailed-description task where we report hallucination and captioning metrics and conduct a GPT4-V assisted evaluation.

\paragraph {Polling-based Object-Probing Evaluation} POPE consists of  object-presence binary questions on 500 COCO dataset images \citep{lin2015microsoft}, with questions equally divided between present and absent objects. Absent objects are chosen based on three criteria: \textit{random}, \textit{popular} (common in COCO), and \textit{adversarial} (commonly co-occurring with present objects). POPE's drawback is its one-word response structure, which limits the influence of decoding strategies and does not evaluate open-ended generation capabilities.

\paragraph{Image Detailed-Descriptions} To complement POPE, we introduce a long-form text generation task called "Image Detailed-Descriptions," inspired by findings from \citet{zhou2023analyzing} that more extensive context increases the likelihood of hallucinations. In this task, the input consists of an image from the COCO dataset and a text prompt requesting a detailed description of the image. The expected output is a long-form, detailed textual description of the given image, typically containing multiple sentences. The prompts used in this task are detailed in appendix~\ref{appendix:exp_details}. By using the same COCO images as POPE, we maintain consistency in the visual domain while exploring LCD's effectiveness in a more challenging setting where the model is required to generate longer, more descriptive outputs.

\paragraph{Baselines and Metrics}

For POPE, we use sampling as the baseline and report F1 scores.\footnote{Complete POPE results are in the appendix, table \ref{table:complete_pope_results}} For the detailed-descriptions task, we use as a baseline  the popular nucleus sampling algorithm\footnote{We find that nucleus-sampling gives better results than vanilla sampling (see table \ref{table:description_results_ablations} in the appendix for ablations).} and report CHAIR metrics \citep{objectHallucination}. To assess description quality, we use captioning metrics against COCO's gold captions, which serve as an approximation considering length differences. Additionally, following \citet{yin2023woodpecker}, we use GPT4-V to evaluate the descriptions for Detailedness and Accuracy (see details in Appendix \ref{appendix:exp_details}).

\paragraph{Models}
We conduct our experiments with leading LVLMs: two versions of the InstructBLIP model (with Flan-T5 and Vicuna LLMs), LLAVA 1.5 and mPLUG-Owl2. The complete experimental details, such as exact model variants and generation hyper-parameters, are given in the Appendix.

\section{Results and Discussion}
For the POPE task, which evaluates object hallucinations using binary questions, LCD improves F1 scores across 11 out of 12 configurations compared to the baseline (Table \ref{table:pope_results}). This suggests that LCD is effective in reducing object hallucinations in the POPE setting. It is worth noting that the POPE setting is highly constrained for decoding algorithms, as it consists of binary yes/no questions, and typically involves only a single decoding step. This limits the potential impact of decoding strategies on the model's performance in this specific task.

In the detailed-description task, which involves generating detailed descriptions of images, LCD significantly reduces hallucinations at both sentence and instance levels across all four models tested (Table \ref{table:description_results}). However, it is important to note that despite the improvements, the CHAIR scores, which measure hallucination rates (lower is better), remain relatively high. This indicates that object hallucinations are still prevalent in long-form LVLM outputs, even with the application of LCD.\footnote{Examples of generated descriptions are found in Appendix \ref{appendix:detailed_descriptions_images}}

We observe that LCD is particularly effective in improving the performance of InstructBLIP models (InstructBLIP$_\text{F}$
and InstructBLIP$_\text{V}$). We hypothesize that this may be due to the fact that the LLMs in these models are frozen during training, which results in a stronger language bias that LCD can effectively mitigate.
When evaluating the overall generation quality using captioning metrics (METEOR, WMD, and ROUGE$_\text{L}$), LCD outperforms the baseline in all cases except one (WMD in LLAVA 1.5, where the reduction is approximately 1\%). This indicates that LCD not only reduces hallucinations but also maintains or improves the overall quality of the generated descriptions.

Furthermore, in the GPT4-V assisted evaluation, which assesses the accuracy and detailedness of the generated descriptions, LCD improves the accuracy scores across all models. Interestingly, the detailedness scores remain similar to the baseline, suggesting that LCD reduces hallucinations without increasing the granularity of the descriptions.

\begin{table}[t]
\centering
\resizebox{\columnwidth}{!}{
\begin{tabular}{l|l||l|l}
\hline
\textbf{POPE} & \textbf{Model} & \textbf{Baseline F1} & \textbf{LCD F1} \\ \hline
Random &  & 83.95 & \textbf{87.55} \\
Popular & InstructBLIP$_\text{V}$  & 82.80 & \textbf{84.34} \\
Adversarial &  & 80.25 & \textbf{81.64} \\
 \hline
Random & & 84.05 & \textbf{84.27} \\
Popular & InstructBLIP$_\text{F}$  &  80.74 & \textbf{82.81} \\
Adversarial &  & 78.87 & \textbf{80.69} \\
\hline
Random &  & \textbf{84.17} & 83.76 \\
Popular & LLAVA 1.5 & 83.10 & \textbf{83.47} \\
Adversarial &  & 81.34 & \textbf{81.62} \\
\hline
Random &  & 86.96 & \textbf{87.51} \\
Popular & mPLUG-Owl2  & 82.88 & \textbf{84.93} \\
Adversarial &  & 82.93 & \textbf{83.91} \\
\hline
\end{tabular}
}
\caption{POPE results for different models and methods.}
\label{table:pope_results}
\end{table}

\section{Conclusion}
In this paper we present Language Contrastive Decoding, a novel method to reduce hallucinations in LVLMs. By dynamically adjusting output probabilities using the LVLM's internal LLM, LCD significantly improves hallucination metrics across different LVLM architectures, enhancing the quality and reliability of generated content without necessitating retraining or auxiliary models and post-processing. This work highlights the potential of specialized decoding strategies in enhancing multimodal AI models and lays the groundwork for further exploration into more sophisticated LVLM decoding methods.

\section{Limitations}

Firstly, while LCD shows promise in reducing hallucinations, it only targets hallucinations caused by language biases, but hallucinations can arise from other sources. For instance, previous work has shown that some hallucinations are caused by poor visual understanding \citep{guan2023hallusionbench}. We believe LCD can be used as a platform to craft LVLM-specific decoding algorithms that would mitigate hallucinations stemming from different factors, and leave this pursuit for future work.

Secondly, our evaluation method primarily addresses object hallucinations, which are only one form of hallucination that LVLMs may exhibit. Preliminary results signal that LCD mitigates more complex manifestations of language-induced hallucinations as assessed by recent benchmarks such as FAITHSCORE \citep{faithscore} and HallusionBench \citep{guan2023hallusionbench}, but further work is required to establish this.

Moreover, LCD relies on current LVLM architectures that combine an LLM and a text-vision encoder, and requires access to an LLM that emits output probabilities on the same set of tokens as the LVLM. It is possible that the future generation of multimodal AI systems will have a different architecture that will make LCD obsolete. Additionally, LCD requires an LLM forward pass for each LVLM decoding step. The added latency could be mitigated with efficient inference techniques, and also by using a smaller LLM as the contrasting model. The effectiveness of LCD in this scenario is left for future work.

Finally, there are ethical considerations related to the mitigation of hallucinations in LVLMs. As these models become more reliable, it is crucial to continue evaluating the potential impacts of their use, ensuring they do not perpetuate or exacerbate biases present in their training data. LCD indeed mitigates some biases, but it is important to keep in mind that it might amplify other biases, unknown to us. Responsible deployment of these models requires ongoing vigilance and a commitment to transparency and fairness.

\section*{Acknowledgements}

We  thank Yoav Goldberg, Ido Dagan, and the participants of the NLP seminar at Bar-Ilan University for their valuable feedback.
This research has been funded by a grant from
the European Research Council, ERC-StG grant
number 677352, and a grant by the Israeli  Science Foundation (ISF), grant number 670/23, for which we are grateful. 

\bibliography{pruned_references}

\appendix

\onecolumn
\section{Appendix}
\label{sec:appendix}

\subsection{Detailed Experimental Setup}
\label{appendix:exp_details}

For POPE and the descriptions experiment, we use the following LCD parameters $\beta=3.0$, $\alpha=0.1$. 
We set the temperature to $0.5$ in POPE and $1.0$ in the descriptions experiment.
We limit the descriptions length to 250 tokens in all models we tested.
We don't tune any of these parameters.
The prompt we use for the descriptions experiment is \textit{"Describe this image in detail:"}.
The models we use have the following Huggingface identifiers:
\begin{itemize}
\item{Salesforce/instructblip-vicuna-7b} \item{Salesforce/instructblip-flan-t5-xl} \item{llava-hf/llava-1.5-7b-hf}
\item{MAGAer13/mplug-owl2-llama2-7b}
\end{itemize}

\begin{figure}[!htb]
\centering
\includegraphics[scale=0.38]{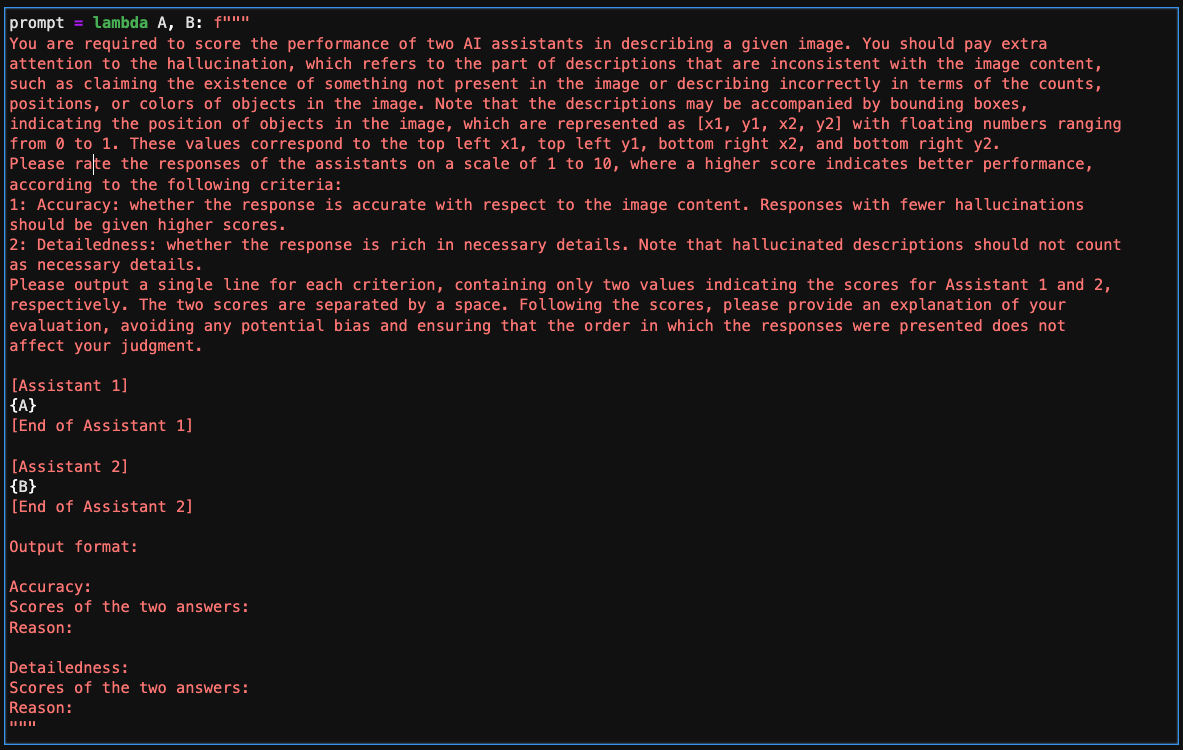}
\parbox{\textwidth}{\caption{Prompt used to evaluate descriptions with GPT4-V, taken from \citet{yin2023woodpecker}}
\label{fig:gpt4_prompt}
}

\end{figure}

\paragraph{GPT4-V Assisted Evaluation}
We follow the evaluation protocol given in \citet{yin2023woodpecker}, where an image and two descriptions are given to the model, formatted with the prompt in figure \ref{fig:gpt4_prompt}. The model outputs scores in two dimensions: Accuracy and Detailedness. We use the \textit{gpt-4-vision-preview} model on February 2024.

\clearpage

\subsection{COCO Detailed Descriptions Examples}
The descriptions in this section were generated by the LLAVA 1.5 model.
\label{appendix:detailed_descriptions_images}
\begin{figure}[!htpb]
\captionsetup{font=footnotesize, justification=raggedright, singlelinecheck=false}

\includegraphics[width=.5\linewidth]{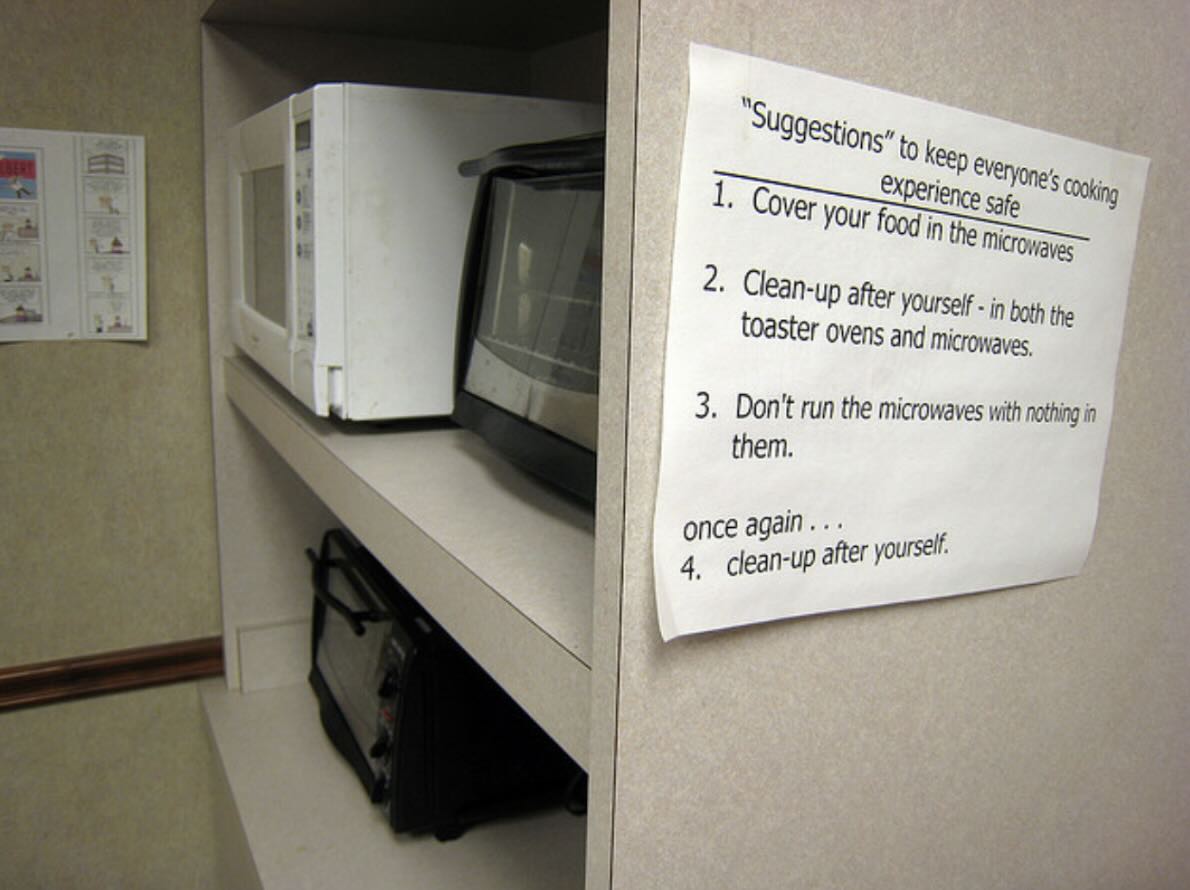}

\caption*{COCO Image 461331}

\scriptsize
\textbf{Nucleus Sampling:} The image features a white counter area with various appliances and a sign. On top of the counter, there is a microwave positioned alongside a toaster oven. Below the toaster oven is a sign with suggestions to keep people safe while using the kitchen. Several appliances are present, including two cups, one being closer to the foreground, and the other near the edge of the counter. A sink is located near the lower left corner, and a spoon can be seen towards the right side of the counter. There are two ovens, with one situated towards the right side, and the other near the center of the counter. Lastly, a bottle is visible in the background, at the far left edge.

\textbf{LCD:} The image showcases a kitchen setting featuring a microwave sitting on top of a shelf. Below the microwave, a toaster oven is placed in close proximity. There are also a couple of toaster ovens situated further back in the scene. A note is posted on the shelf, offering suggestions to keep everyone safe by covering one's experience while using the appliances, cleaning up after themselves, and not turning on the microwaves with no one around. The scene reflects a cooperative environment for kitchen users.
\vspace{0.5cm}

\captionsetup{font=footnotesize, justification=raggedright, singlelinecheck=false}

\includegraphics[width=.5\linewidth]{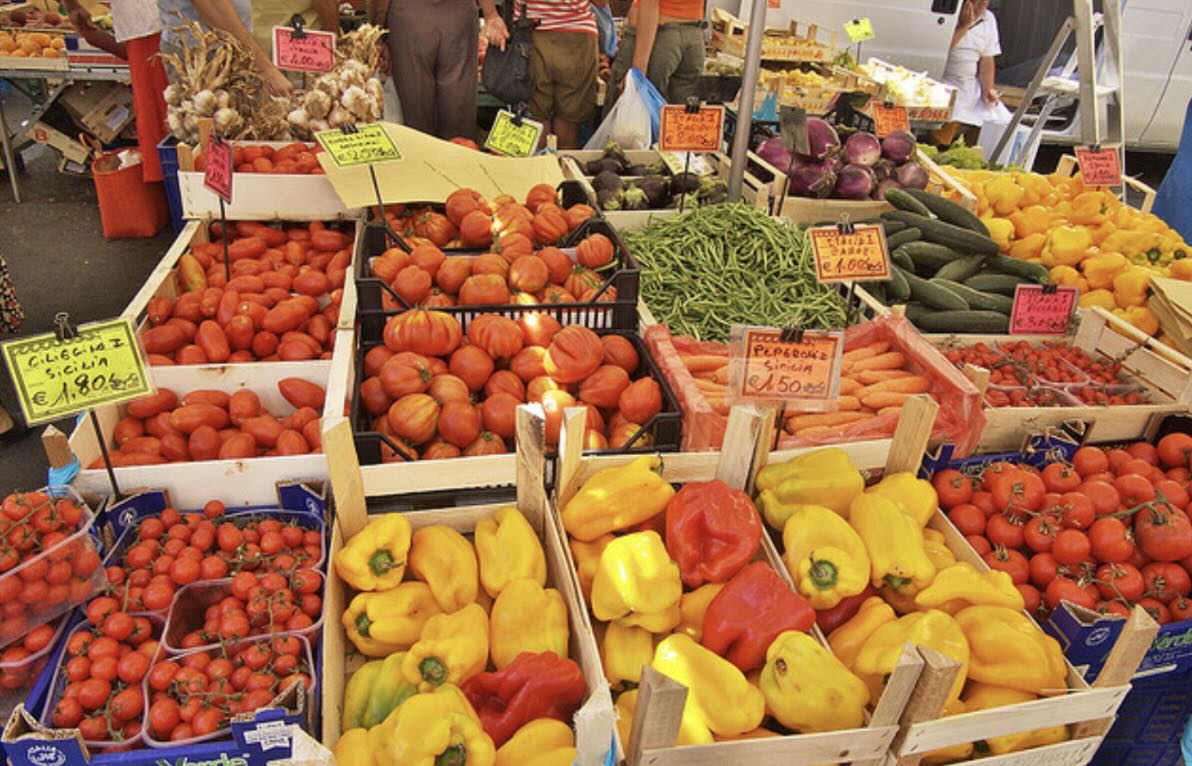}
\caption*{COCO Image 378873}

\textbf{Nucleus Sampling:} The image features an outdoor market with a variety of fruits and vegetables for sale. Two people can be seen in the scene, likely browsing the options and making their selections. The produce consists of several bunches of ripe red tomatoes, and smaller, freshly picked green peppers. There is also a larger selection of yellow bell peppers available in various sizes. In addition to the vegetables, there is a bottle visible near the top left corner. Overall, the market offers an array of fresh, colorful, and healthy products for customers to choose from.

\textbf{LCD:} The image showcases an outdoor produce market filled with an abundant selection of fruits and vegetables. Numerous tomatoes can be found throughout the market, in various locations and sizes. Peppers are another prominent feature at this market, with multiple bunches spread across different parts of the display. A person is visible near the top left of the scene, likely browsing through the market offerings or engaging with the sellers. Overall, it's a lively marketplace filled with fresh and healthy options.
\end{figure}

\begin{figure}[!ht]
\captionsetup{font=footnotesize, justification=raggedright, singlelinecheck=false}

\includegraphics[width=.3\linewidth]{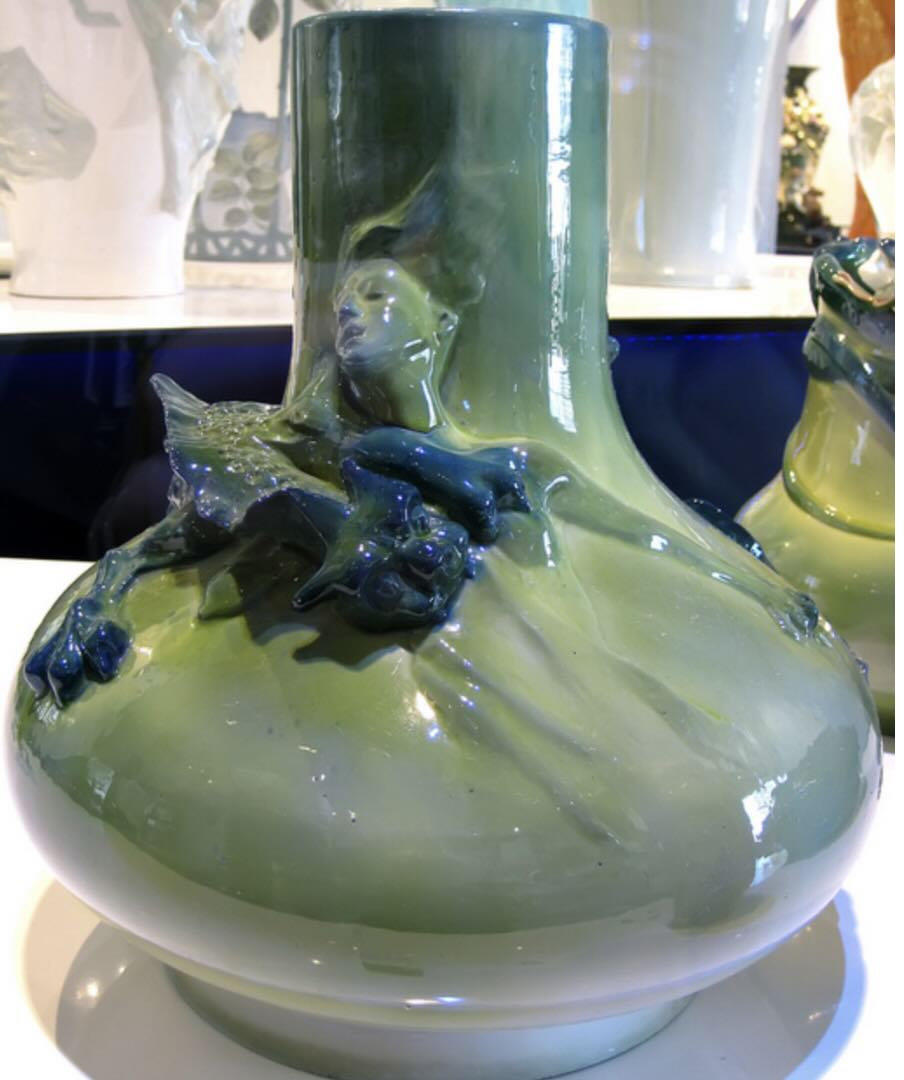}
\caption*{COCO Image 498100}

\scriptsize
\textbf{Nucleus Sampling:} The image displays a room filled with numerous decorative items. A prominent green ceramic vase stands out, with a colorful art deco sculpture depicting a woman laying on it. There are also two smaller vases in the scene – one in the upper right corner and the other closer to the center of the room. The room is adorned with a variety of objects, including a clock on the wall, a collection of books scattered in the background, and a chair positioned in the middle. This visually rich space showcases a diverse assortment of vases, sculptures, and other home decor items.

\textbf{LCD:} The image showcases an assortment of vases on display, featuring a variety of colors and sizes. Among these, there is a large green vase as the focal point of the scene. Inside this large green vase, a statue of a woman can be seen lying on the side of the vase. Another green vase is located towards the right side of the scene. Additionally, a smaller green vase is found close to the center. Two other smaller vases can also be spotted within the collection. The diverse display of vases creates an engaging and artistic atmosphere for visitors to enjoy.

\vspace{0.5cm}

\includegraphics[width=.5\linewidth]{images/image4.jpg}
\caption*{COCO Image 312406}

\scriptsize
\textbf{Nucleus Sampling:} The image features a tray of cake covered with foil and a plastic container with coconut powder, all placed on a counter. There are several sliced chocolate cakes, some of which are stacked. Additionally, a pair of tongs is resting beside the cakes, ready to be used for serving. A bowl with a spoon in it is also situated on the table, accompanied by a banana and a spoon nearby. The overall arrangement looks like a setting for a delicious dessert.

\textbf{LCD:} The image features a dining table with various items arranged on it. A banana can be seen prominently, occupying a large portion of the left side of the table. Along with the banana, there's a can of powdered cocoa placed towards the top-middle area of the table. Additionally, two bowls are present on the table – one occupying the middle-right part and another covering a smaller section on the right side. There is also a spoon situated on the right side of the table, ready for use in enjoying the snack or dessert.
\end{figure}

\begin{figure}[h]
\captionsetup{font=footnotesize, justification=raggedright, singlelinecheck=false}

\includegraphics[width=.5\linewidth]{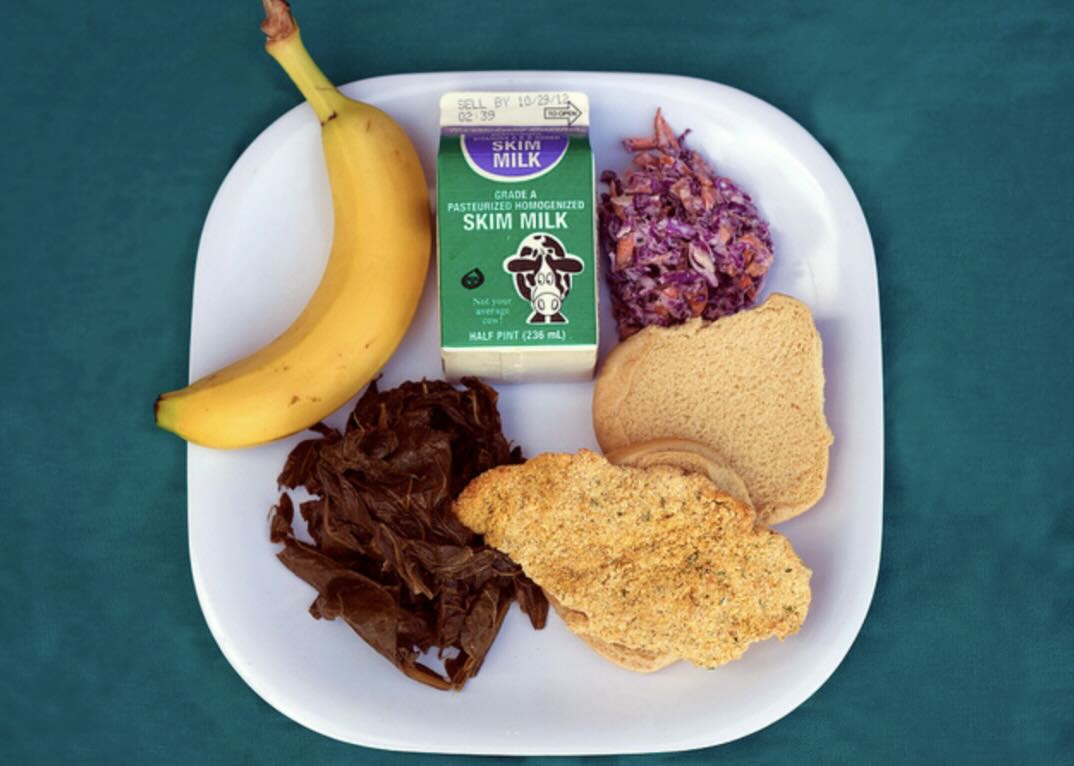}
\caption*{COCO Image 379404}

\scriptsize
\textbf{Nucleus Sampling:} The image displays a delicious meal consisting of chicken, broccoli, and a banana. The plate is large and fills most of the frame, with various items on the plate. The chicken is located at the bottom-right side of the plate, while the broccoli pieces are scattered throughout the plate. Additionally, there is a banana on the plate that is placed towards the top left corner. In the background, a bottle can be seen, likely containing a drink that complements the meal.

\textbf{LCD:} The image features a white plate topped with various foods. On the plate, there's a banana positioned towards the left, accompanied by some meat, likely fish sticks. Alongside the main items, a glass of skim milk is present, adding to the nutritious meal. There is also a sandwich, possibly a piece of bread, contributing to the diverse meal displayed on the plate.

\vspace{0.5cm}

\includegraphics[width=.5\linewidth]{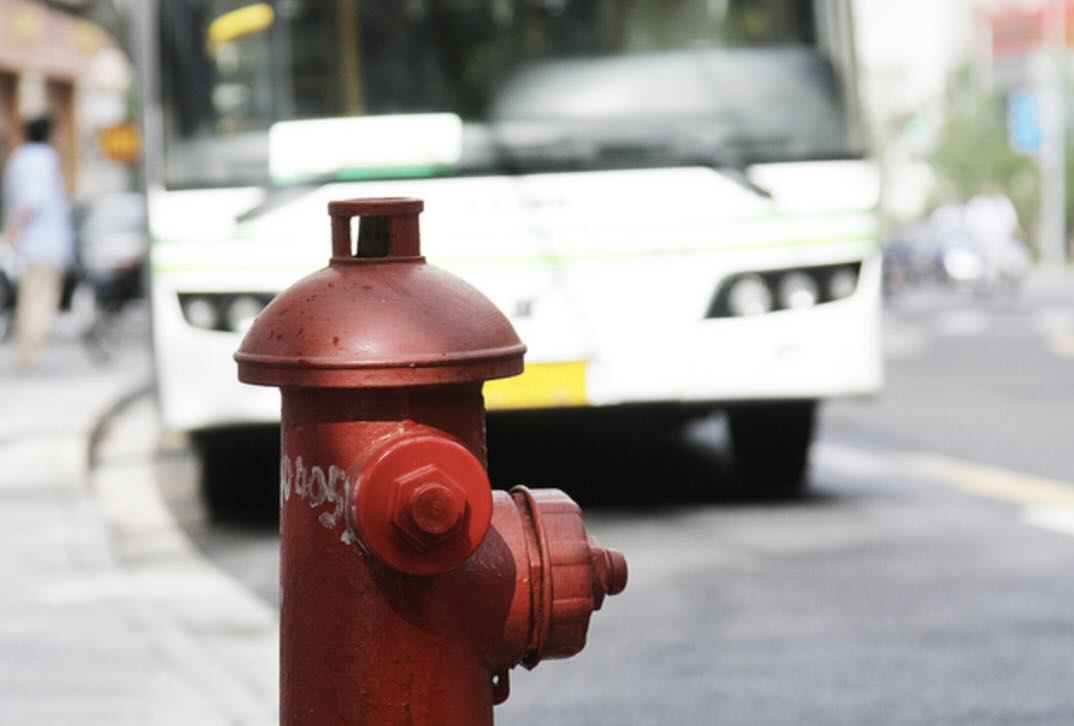}
\caption*{COCO Image 125524}

\scriptsize
\textbf{Nucleus Sampling:} The image features a bright red fire hydrant prominently placed on the side of the street. The fire hydrant is situated close to a bus stop and appears to be unused. A large white bus is positioned behind the fire hydrant on the other side of the street. A car is also visible in the background, slightly to the right. In addition, there are two people in the scene, both located at the far right side. One of them is holding a cell phone, possibly capturing the moment or checking their phone.

\textbf{LCD:} In the scene, a red fire hydrant is situated next to the street, capturing attention due to its striking color. A bus is driving down the road nearby, positioned closer to the background of the image. There are several people dispersed along the street. Some of them can be seen closer to the middle and right side of the scene, while others are scattered across the entire street, possibly going about their day or waiting to board the bus.
\end{figure}

\clearpage

\subsection{Detailed Experimental Results}
\label{appendix:detailed_results}

\begin{table}[!hbt]
\centering
\begin{tabular}{lllllll}
\hline
\textbf{Model} & \textbf{Method} & \textbf{METEOR}$\uparrow$ & \textbf{WMD}$\uparrow$ & \textbf{ROUGE$_\text{L}$}$\uparrow$ & \textbf{CHAIRs}$\downarrow$ & \textbf{CHAIRi}$\downarrow$ \\ \hline
InstructBLIP$_\text{F}$ & Baseline & 0.151 & 0.361 & 0.156 & 0.666 & 0.174 \\
 & Baseline$_\text{N}$ & 0.157 & \underline{0.367} & 0.161 & 0.662 & 0.146 \\

 & LCD$_\text{-dw}$ & \underline{0.159} & 0.364 & \underline{0.163} & \underline{0.594} & \underline{0.133} \\
 & LCD & \textbf{0.163} & \textbf{0.370} & \textbf{0.168} & \textbf{0.566} & \textbf{0.131} \\
\hline
InstructBLIP$_\text{V}$ & Baseline & 0.171 &	0.408 &	0.274 &	0.308 & 0.138 \\
& Baseline$_\text{N}$ & 0.178 & 0.423 & 0.291 & 0.274 & 0.126 \\
 & LCD$_\text{-dw}$ & \textbf{0.202} & \underline{0.474} & \underline{0.366} & \underline{0.23} & \underline{0.116} \\
 & LCD & \underline{0.199} & \textbf{0.48} & \textbf{0.38} & \textbf{0.174} & \textbf{0.107} \\
\hline
LLAVA 1.5 & Baseline & 0.160 & \underline{0.353} & 0.167 & 0.632 & 0.183 \\
& Baseline$_\text{N}$ & 0.163 & \textbf{0.357} & 0.169 & 0.672 & 0.182 \\
 & LCD$_\text{-dw}$ & \underline{0.169} & 0.352 & \underline{0.179} & \underline{0.624} & \textbf{0.157} \\
 & LCD & \textbf{0.171} & 0.352 & \textbf{0.181} & \textbf{0.610} & \underline{0.161} \\
 \hline
\end{tabular}
\parbox{\textwidth}{\caption{Image Description ablations. \textit{-dw} is an LCD variant without dynamic weighting, with $\beta=0.5$. Baseline$_\text{N}$ is using nucleus sampling with $p =0.95$, Baseline is vanilla sampling.
}
\label{table:description_results_ablations}
}
\end{table}

\label{appendix:complete_POPE}

\begin{table}[!hbt]
\centering
\begin{adjustbox}{max width=\textwidth}
\begin{tabular}{@{}l|l|l|r|r|r|r|r@{}}
\toprule
\textbf{POPE} & \textbf{method} & \textbf{model}      & \multicolumn{1}{l|}{\textbf{accuracy}} & \multicolumn{1}{l|}{\textbf{precision}} & \multicolumn{1}{l|}{\textbf{recall}} & \multicolumn{1}{l|}{\textbf{f1}} & \multicolumn{1}{l}{\textbf{yes ratio}} \\ \midrule
random        & Baseline        & InstructBLIP Vicuna & 84.90\%                                & 89.57\%                                 & 79.00\%                              & 83.95\%                          & 44.10\%                                 \\
random        & LCD             & InstructBLIP Vicuna & 87.53\%                                & 87.43\%                                 & 87.67\%                              & 87.55\%                          & 50.13\%                                 \\
popular       & Baseline        & InstructBLIP Vicuna & 83.30\%                                & 85.35\%                                 & 80.40\%                              & 82.80\%                          & 47.10\%                                 \\
popular       & LCD             & InstructBLIP Vicuna & 83.73\%                                & 81.31\%                                 & 87.60\%                              & 84.34\%                          & 53.87\%                                 \\
adversarial   & Baseline        & InstructBLIP Vicuna & 80.23\%                                & 80.17\%                                 & 80.33\%                              & 80.25\%                          & 50.10\%                                 \\
adversarial   & LCD             & InstructBLIP Vicuna & 80.27\%                                & 76.33\%                                 & 87.73\%                              & 81.64\%                          & 57.47\%                                 \\
random        & Baseline        & InstructBLIP FlanT5 & 85.63\%                                & 94.43\%                                 & 75.73\%                              & 84.05\%                          & 40.10\%                                 \\
random        & LCD             & InstructBLIP FlanT5 & 86.03\%                                & 96.47\%                                 & 74.80\%                              & 84.27\%                          & 38.77\%                                 \\
popular       & Baseline        & InstructBLIP FlanT5 & 82.07\%                                & 87.17\%                                 & 75.20\%                              & 80.74\%                          & 43.13\%                                 \\
popular       & LCD             & InstructBLIP FlanT5 & 84.43\%                                & 92.44\%                                 & 75.00\%                              & 82.81\%                          & 40.57\%                                 \\
adversarial   & Baseline        & InstructBLIP FlanT5 & 79.83\%                                & 82.83\%                                 & 75.27\%                              & 78.87\%                          & 45.43\%                                 \\
adversarial   & LCD             & InstructBLIP FlanT5 & 82.03\%                                & 87.22\%                                 & 75.07\%                              & 80.69\%                          & 43.03\%                                 \\
random        & Baseline        & LLAVA 1.5           & 85.87\%                                & 95.67\%                                 & 75.13\%                              & 84.17\%                          & 39.27\%                                 \\
random        & LCD             & LLAVA 1.5           & 85.73\%                                & 97.18\%                                 & 73.60\%                              & 83.76\%                          & 37.87\%                                 \\
popular       & Baseline        & LLAVA 1.5           & 84.80\%                                & 93.57\%                                 & 74.73\%                              & 83.10\%                          & 39.93\%                                 \\
popular       & LCD             & LLAVA 1.5           & 85.40\%                                & 96.17\%                                 & 73.73\%                              & 83.47\%                          & 38.33\%                                 \\
adversarial   & Baseline        & LLAVA 1.5           & 82.77\%                                & 88.67\%                                 & 75.13\%                              & 81.34\%                          & 42.37\%                                 \\
adversarial   & LCD             & LLAVA 1.5           & 83.33\%                                & 90.98\%                                 & 74.00\%                              & 81.62\%                          & 40.67\%                                 \\ \bottomrule
\end{tabular}
\end{adjustbox}
\parbox{\textwidth}{\caption{Complete POPE results.
}
\label{table:complete_pope_results}
}
\end{table}

\end{document}